\documentclass{article}

\usepackage[main, final]{neurips_2026}

\usepackage[utf8]{inputenc} % allow utf-8 input
\usepackage[T1]{fontenc}    % use 8-bit T1 fonts
\usepackage{hyperref}       % hyperlinks
\usepackage{url}            % simple URL typesetting
\usepackage{booktabs}       % professional-quality tables
\usepackage{amsfonts}       % blackboard math symbols
\usepackage{amsmath}
\usepackage{nicefrac}       % compact symbols for 1/2, etc.
\usepackage{microtype}      % microtypography
\usepackage{xcolor}         % colors
\usepackage{graphicx}
\usepackage{booktabs}   % 提供三线表命令
\usepackage{graphicx}   % 提供缩放功能
\usepackage{calc}       % 提供计算功能
\usepackage{caption}      % added by gjx
\usepackage{wrapfig}
\usepackage{booktabs}
\usepackage{subcaption} % 必须引入这个宏包
\usepackage{graphicx}
\usepackage{xurl}
\newcommand{\method}{\textit{D-VLA }}
\title{D-VLA: A High-Concurrency Distributed Asynchronous Reinforcement Learning Framework for Vision-Language-Action Models }

\author{%
  Yongjian Guo\textsuperscript{1,3}\thanks{Equal contributions.} \quad
  Yucheng Guo\textsuperscript{3}\footnotemark[1] \ \thanks{Corresponding author.} \quad
  Zhong Guan\textsuperscript{2,3} \quad
  Wen Huang\textsuperscript{1,3} \\
  \bfseries Haodong Yue\textsuperscript{1} \quad
  Shuai Di\textsuperscript{3} \quad
  Xiong Jun Wu\textsuperscript{3} \quad
  Yicheng Gong\textsuperscript{3} \quad
  Xi Xiao\textsuperscript{1} \footnotemark[2] \\
  \textsuperscript{1}Tsinghua Shenzhen International Graduate School, Tsinghua University \\
  \textsuperscript{2}Tianjin University \quad
  \textsuperscript{3}JDT AI Infra \\
}
\begin{document}

\maketitle
% Unnumbered code link: preserve the author footnote symbols.
\begingroup
\renewcommand{\thefootnote}{}
\footnotetext{\Urlmuskip=0mu plus 1mu\relax
  The source code is available at \url{https://github.com/Aurora24Thu/D-VLA}.}

\endgroup

\begin{abstract}
Reinforcement learning (RL) for Vision-Language-Action (VLA) models is constrained by resource contention between physics simulation and model computation, as well as communication and scheduling overhead. We present \method, a distributed asynchronous RL framework that separates high-frequency trajectory transfer over NCCL from lower-frequency weight synchronization over a CPU-side Gloo channel. A four-thread ``Swimlane'' pipeline overlaps rollout, weight reception, training, and parameter distribution, while dual-pool GPU memory management and an in-process zero-copy observation path support high-concurrency simulation. We evaluate $\pi_{0.5}$ and OpenVLA-OFT on GPU-intensive ManiSkill and CPU-intensive LIBERO. Mechanism-level experiments isolate improvements in synchronization stability, memory capacity, and observation-transfer latency. On LIBERO, asynchronous scheduling improves throughput despite the absence of the same GPU-physics contention. We further evaluate both GRPO and PPO on ManiSkill. In a separate scaling study with $\pi_{0.5}$, \method retains 86.70\% throughput scaling efficiency from 8 to 256 GPUs. 
%The data is available in https://anonymous.4open.science/r/D-VLA-F3BD/README.md
\end{abstract}

\section{Introduction}
\begin{wrapfigure}{r}{0.5\linewidth}
  \vspace{-0.8em}
  \includegraphics[width=0.85\linewidth]{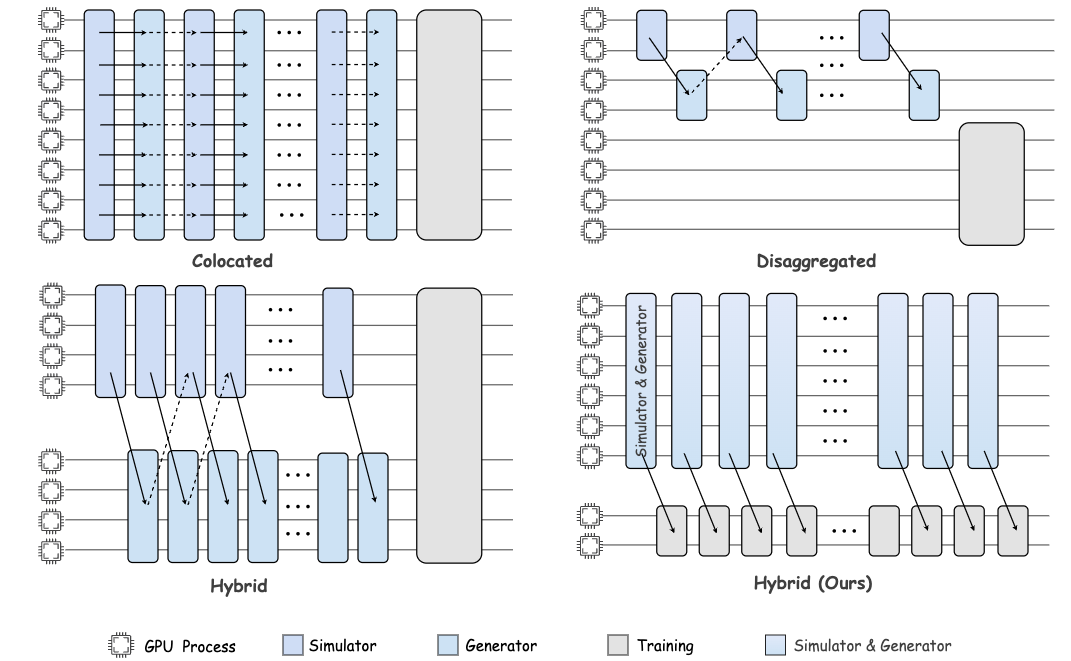}
  \caption{Placement Strategies across Different Training Frameworks}
  \label{fig:preview}
  \vspace{-0.8em}
\end{wrapfigure}

Embodied AI, regarded as a pivotal pathway toward Artificial General Intelligence (AGI), is undergoing a profound paradigm shift driven by the emergence of Vision-Language-Action (VLA) models \cite{zitkovich2023rt2, black2024pi_0, gemini2025robotics, kim2024openvla, shukor2025smolvla} such as OpenVLA \cite{kim2024openvla}, $\pi_0$ \cite{black2024pi_0}, and GR00T \cite{bjorck2025gr00t}. These models achieve a significant transition from manually designed explicit models to data-driven implicit models by integrating visual perception, language understanding, and action generation into a unified end-to-end framework \cite{ma2024survey, zhong2025survey, zhang2025pure}. Through continuous scaling across massive computational resources and datasets, VLA models have demonstrated unprecedented potential in cross-task and cross-morphology adaptation. However, despite these notable advancements, current training paradigms remain heavily reliant on imitation learning based on Supervised Fine-Tuning (SFT). Existing frameworks like LeRobot \cite{cadene2026lerobot} and GR00T \cite{bjorck2025gr00t} primarily utilize expert demonstration data to fine-tune policies via behavior cloning. This SFT-centric path faces multiple severe challenges in practical applications: first, large-scale human-collected robot trajectory data is both costly and difficult to obtain, which strictly limits the scaling of models \cite{sheng2025hybridflow, hu2024openrlhf}; second, processing high-dimensional data and complex multimodal architectures imposes a heavy burden on training cycles and inference latency. Furthermore, constrained by the limited diversity of offline datasets, SFT models often exhibit weak generalization capabilities when encountering distribution shifts and unseen tasks \cite{chen2025minimax, su2025klear}. Unlike human learning mechanisms, SFT cannot support agents in discovering new action patterns through autonomous exploration beyond demonstration data. Consequently, the research community is increasingly pivoting toward Reinforcement Learning (RL) frameworks to break through these limitations via online interaction.

In response to the deficiencies of SFT, several RL frameworks have emerged, optimizing different dimensions of the training pipeline. For instance, RLinf \cite{zang2025rlinf} provides a universal distributed training and evaluation interface, offering a unified framework for multimodal agents and VLAs. RL-VLA$^3$ \cite{guan2026rl} implements a three-stage asynchronous pipeline to decouple data collection, policy inference, and model updates, thereby maximizing hardware utilization and training throughput. SimpleVLA-RL \cite{li2025simplevla} designs rule-based outcome rewards and interactive trajectory sampling for VLA models, proving that performance can surpass SFT even with minimal demonstrations. Additionally, Dexbotix \cite{xie2025dexbotic} focuses on integrating tactile feedback for high-degree-of-freedom dexterous hands, while Vlab \cite{aubakirova2025vlab} is dedicated to building specialized sim-to-real transfer environments.

Unlike traditional online RL, Embodied AI training involves deep coupling between high-fidelity physics simulation and large-scale deep learning models. The former features high-frequency, fragmented occupation of computational resources, while the latter demands extreme throughput for GPU memory capacity and communication bandwidth. Existing distributed training frameworks, such as RLinf-VLA \cite{zang2025rlinf} and RL-VLA$^3$ \cite{guan2026rl}, introduce hybrid resource allocation and fine-grained asynchronous mechanisms to alleviate computational pressure. However, they have yet to fundamentally resolve the resource contention and execution conflicts between simulation tasks and model optimization at the underlying architecture level. As a result, the overall system throughput remains bottlenecked by the slowest physics stepping or synchronization overhead.

The current performance bottlenecks in embodied RL systems primarily stem from the high degree of coupling between simulation logic and learning logic on the execution plane. On one hand, frequent memory allocation and deallocation by physics engines easily lead to severe memory fragmentation in deep learning frameworks. On the other hand, the frequent transfer of massive multimodal environment data (e.g., high-resolution images) between sampling and inference components introduces significant serialization overhead and communication latency. This systemic "blocking" effect is particularly severe when handling long-sequence interaction tasks, restricting the sample acquisition efficiency of agents in complex scenarios.

To address these challenges, we propose \method, a \textbf{d}istributed \textbf{VLA} RL framework. Its ``Plane Decoupling'' design separates high-frequency trajectory communication from lower-frequency weight synchronization to reduce interference with simulation. A four-thread ``Swimlane'' pipeline overlaps sampling, weight reception, gradient training, and parameter distribution; the achievable overlap depends on the balance between rollout and actor workloads. Dual-pool memory management and a zero-copy observation path support co-located, separated, and hybrid placements (Figure~\ref{fig:preview}). Local topology replication preserves the locality of environment--inference communication as GPU count grows. We evaluate the framework with Group Relative Policy Optimization (GRPO)~\cite{shao2024deepseekmath} and Proximal Policy Optimization (PPO), and distinguish the GPU-contention regime of ManiSkill from the host-side scheduling bottlenecks of LIBERO.

The primary contributions of this paper are summarized as follows:
\begin{itemize}
    \item \textbf{Decoupled communication and asynchronous execution:} We separate NCCL trajectory transfer from Gloo weight synchronization and overlap four execution threads, reducing synchronization stalls between simulation and learning.
    \item \textbf{Memory capacity and communication locality:} We combine dual-pool GPU memory management, in-process zero-copy observation access, and local topology replication. Targeted experiments quantify the capacity benefit of memory partitioning and the latency benefit of zero-copy access.
    \item \textbf{Evaluation across workloads and scales:} We benchmark 3B and 7B VLA models on ManiSkill and LIBERO, evaluate GRPO and PPO, and report multi-node weak scaling up to 256 GPUs.
\end{itemize}

\section{Related Work}

\paragraph{Evolution of Vision-Language-Action Models}
In the field of Embodied AI, the evolution of Vision-Language-Action (VLA) models is undergoing a paradigm shift from fundamental Supervised Fine-Tuning (SFT) \cite{zhou2026thousand} toward Reinforcement Learning (RL) \cite{aubakirova2025vlab} frameworks with enhanced generalization capabilities. Early foundational VLA models such as RT-2 \cite{zitkovich2023rt2}, OpenVLA \cite{kim2024openvla}, and $\pi_0$ \cite{black2024pi_0} achieved preliminary robotic manipulation capabilities by training on large-scale datasets like Open-X Embodiment \cite{o2024open}. Subsequently, models like $\pi_{0.5}$ \cite{intelligence2025pi_} and SmolVLA \cite{shukor2025smolvla} further optimized parameter efficiency and inference speed while maintaining performance. Concurrently, general-purpose humanoid control models such as Gr00t N1.5 \cite{bjorck2025gr00t} have demonstrated broad application prospects. However, due to the limitations of SFT methods in handling out-of-distribution data, research focus is gradually shifting toward utilizing RL to enable continuous dynamic environmental adaptation in complex benchmarks such as ManiSkill \cite{mu2021maniskill} and LIBERO \cite{liu2023libero}.

\paragraph{Frameworks for VLA Training and Optimization}
To support the efficient training of large-scale VLAs, the academic community has developed a series of optimization frameworks ranging from low-level control to high-level learning. Frameworks such as LeRobot \cite{cadene2026lerobot} and DexBotix \cite{xie2025dexbotic} provide vertically integrated toolchains for end-to-end learning, covering the entire process from data processing to policy fine-tuning for algorithms like ACT and Diffusion Policy. Early attempts such as SimpleVLA-RL \cite{li2025simplevla} and RLinf-VLA \cite{zang2025rlinf} sought to migrate RLHF workflows from Large Language Models (LLMs) to embodied tasks. Nevertheless, the high latency introduced by physics simulators causes synchronous training pipelines to face severe throughput bottlenecks. To address this, RL-VLA³ \cite{guan2026rl} proposed a fully asynchronous distributed architecture that decouples simulation, inference, and training processes. By drawing on the design philosophies of LLM training systems like veRL \cite{sheng2025hybridflow}, OpenRLHF \cite{hu2024openrlhf}, and ROLLART \cite{gao2025rollart}, it significantly enhances hardware utilization and training efficiency for policy optimization.

\paragraph{System Challenges in Embodied AI}
Embodied AI systems face system-level challenges distinct from those of traditional language models during training, primarily stemming from the uncertainties inherent in physics simulators \cite{zhou2026thousand}. Unlike the relatively stable neural reward models used in LLM training \cite{yu2025reward}, VLA training requires frequent calls to simulation environments like RoboCasa \cite{nasiriany2024robocasa}. These environments exhibit high and fluctuating CPU and GPU resource consumption, which can easily lead to computational idling and resource waste \cite{mu2021maniskill, nasiriany2024robocasa}. To address these pain points, frontier research such as RL-VLA$^3$ \cite{guan2026rl} has introduced dynamic batch scheduling and fine-grained environment sharding techniques, aiming to resolve the asynchronous bottlenecks caused by simulators while maximizing both sample efficiency and system throughput.

\section{System Design of \method}

\begin{figure}
    \centering
    \includegraphics[width=0.8\linewidth]{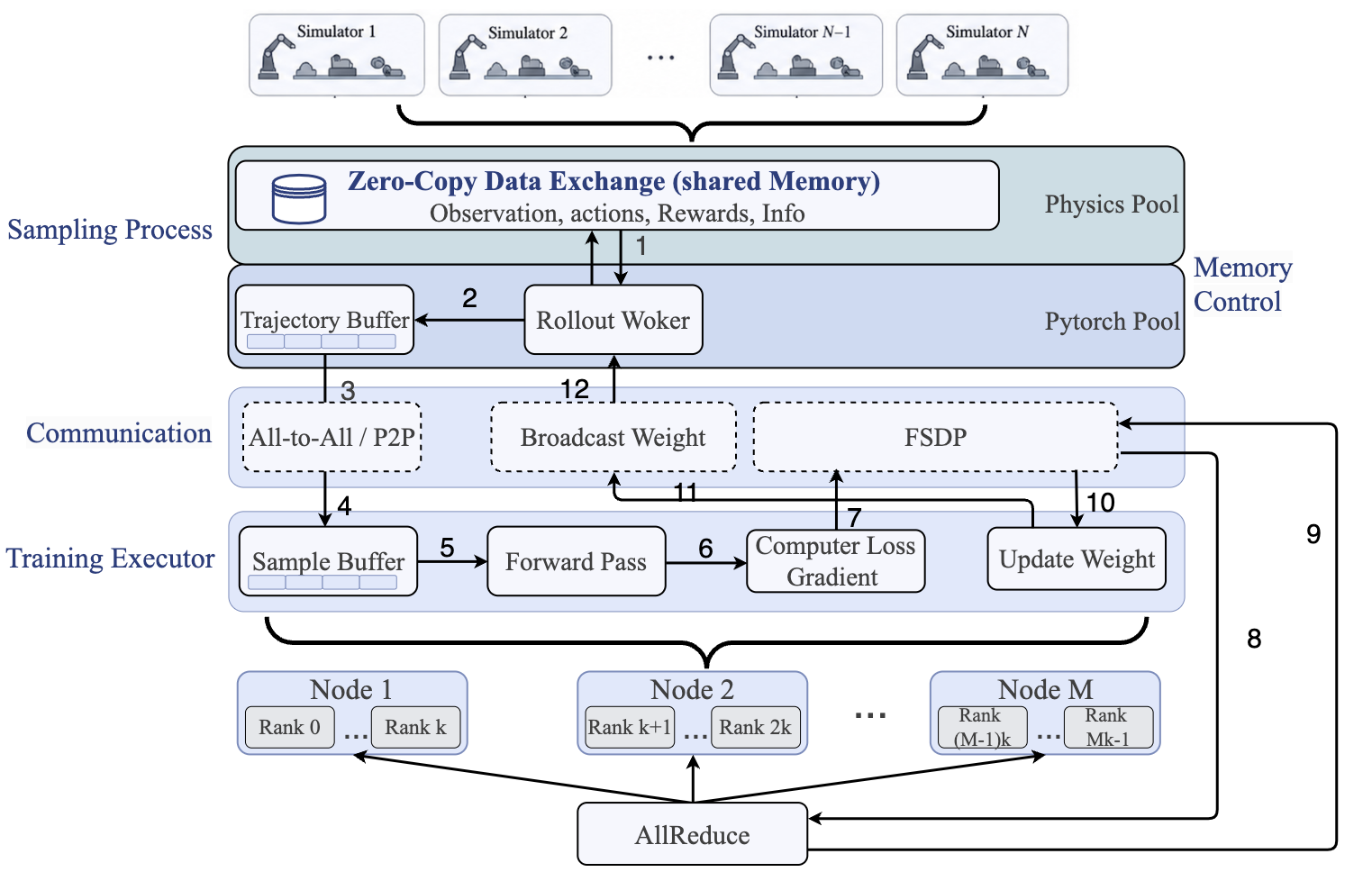}
    \caption{Overview of \method. Rollout GPUs co-locate PhysX environments with an inference policy, enabling in-process observation access. Trajectories are dispatched to actor GPUs over NCCL for training under FSDP; updated weights return through a background Gloo channel. The four-thread pipeline overlaps these stages, with the achievable benefit depending on rollout--actor balance.}
    \label{fig:framework}
\end{figure}

\subsection{\method Framework Overview}
%The rapid advancement of Embodied AI presents severe challenges for the training of Vision-Language-Action (VLA) models, requiring systems to maintain high-frequency interaction with high-fidelity physical environments while processing massive multimodal perceptions. Although existing distributed reinforcement learning frameworks attempt to alleviate computational pressure through asynchronous mechanisms, a fundamental underlying conflict remains between the resource-intensive nature of physical simulation and the high demands of large-scale deep learning models for memory and bandwidth  This often results in system throughput being bottlenecked by performance gaps in specific execution stages 

The \method framework (Figure~\ref{fig:framework}) targets high-concurrency, low-latency VLA training. Its ``Plane Decoupling'' design separates high-frequency data exchange from lower-frequency weight-control logic. This separation reduces interference between communication and simulator synchronization. A four-thread asynchronous pipeline enables overlap between rollout (simulation and inference), weight reception, gradient computation, and parameter distribution.

\subsection{Flexible Deployment and Placement Strategies}
To address different resource requirements, \method supports co-located, separated, and hybrid deployments. Environment simulation involves frequent memory operations, while model inference demands memory bandwidth. Component placement and rollout:actor GPU ratios are fixed for each run according to hardware topology and workload. In hybrid mode, environments and rollout inference share a device group to minimize observation-transfer overhead, while actor training uses a separate group.

To optimize memory overhead, we designed a zero-copy data exchange mechanism for co-located deployments.  Traditional distributed communication involves multiple rounds of data serialization, which creates significant overhead when handling high-resolution images. By maintaining different execution threads within the same process space, \method allows observation results from the simulation environment to be accessed directly by inference components, achieving ``seamless'' data flow at the memory level and reducing bandwidth consumption.  For tasks requiring massive parallel simulation, \method utilizes a separated deployment strategy with pipeline communication, allowing the environment side to perform physical stepping while the inference side concurrently processes the previous decision logic.

\subsection{Decoupled Communication Architecture}
A key component of \method is the separation of the ``Training Data Plane'' from the ``Weight Control Plane.'' Rollout trajectories constitute the bulk data-plane traffic and are transferred using NCCL. Communication patterns are configured according to resource ratios: symmetric placements can use point-to-point transfer, while asymmetric placements distribute batches across actor workers. Metadata and control signaling are kept separate from bulk tensor transfer.

Model weight synchronization uses a CPU-side Gloo backend with contiguous host buffers. Weight updates are less frequent than environment interaction. Moving communication to a background host channel avoids coupling its progress to CUDA synchronization in the PhysX reset path; staging and applying weights still incur data-movement costs. The stability experiment in Section~\ref{sec:ablations} compares the weight-synchronization backends under this scheduling path: the NCCL variant deadlocks, whereas the Gloo configuration completes training.

\subsection{End-to-End Asynchronous Pipeline and Data Flow}

Rollout workers accumulate trajectories and send a batch once the training batch size is reached. An actor-side host-memory queue buffers differences between data production and consumption. Successful episodes are reset and replenished without waiting for a single slow trajectory to finish. For $M$ rollout and $N$ actor workers, the rollout batch is chosen as a multiple of $N$ to distribute training data evenly. These mechanisms reduce sensitivity to individual long episodes.
\begin{wrapfigure}{r}{0.5\linewidth}
  \vspace{-0.8em}
  \includegraphics[width=0.85\linewidth]{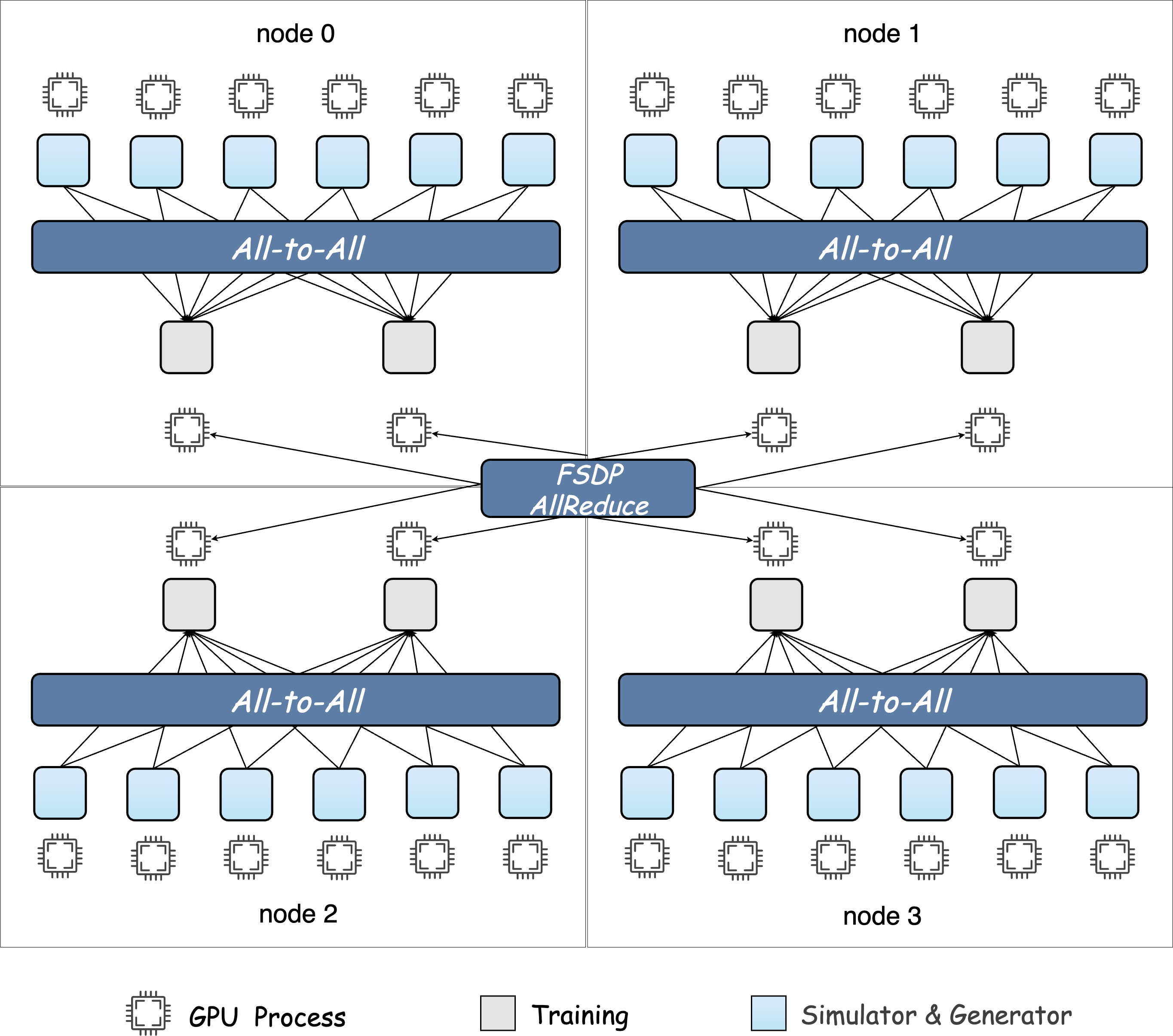}
  \caption{Schematic of Multi-Node Communication in \method}
  \label{multinode}
  \vspace{-0.8em}
\end{wrapfigure}

The ``Swimlane'' model consists of a sampling thread, a weight-receiving thread, a training thread, and a weight-distribution thread, coordinated by lightweight semaphores. Their overlap reduces idle intervals relative to synchronous alternation. The queue interface and communication backend are independent of the policy objective; the evaluated training paths include GRPO with micro-batching and PPO. Episodes have a bounded horizon, limiting trajectory-length variation: the deployment described in our experiments typically needs roughly 40 actions for short successful trajectories and truncates by about 100 actions; the GRPO/PPO comparison specifically uses an 80-step horizon.

\subsection{Memory Management and Large-Scale Node Scaling}
To mitigate memory pressure from non-framework components such as physics engines, \method partitions memory into a ``Model Computation Pool,'' managed by Torch's caching allocator, and an ``Environment Auxiliary Pool'' for physics-engine temporary objects (Figure~\ref{fig:framework}). This partition limits competition between cached model allocations and simulator buffers. It raises the measured concurrency ceiling.

For cross-node scaling, \method replicates a local sampling--inference topology (Figure~\ref{multinode}), preserving the locality of high-frequency observations. In the evaluated topology, environment--rollout traffic remains within each node, while cross-node model synchronization is handled among actor workers using Fully Sharded Data Parallel (FSDP). Background weight distribution separates communication progress from the simulator's CUDA synchronization path. This locality reduces communication overhead as the cluster grows.

\paragraph{Portability.}
The communication and scheduling layers interact with environments through queues carrying observations, actions, and rewards. Adapting a new physics engine therefore requires an environment data adapter rather than a redesign of Plane Decoupling or Swimlane scheduling. Our experiments cover single-agent tasks in ManiSkill and LIBERO.

\section{Experiments}

We evaluate end-to-end throughput, component-level behavior, learning outcomes, and multi-node scaling. ManiSkill stresses GPU simulation and model execution, whereas LIBERO provides a complementary CPU-intensive workload. The system comparison, algorithm comparison, and scaling study use distinct configurations, specified below rather than treated as interchangeable measurements.

\subsection{Experimental Setup}

\begin{figure}[t]
    \centering
    \includegraphics[width=1\linewidth]{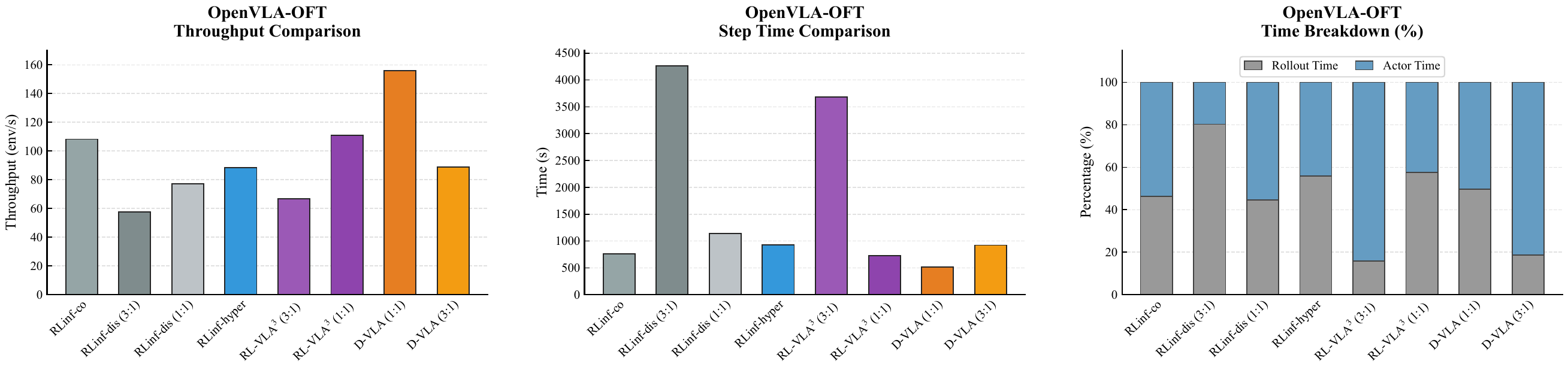}
    \caption{OpenVLA-OFT on ManiSkill with 8 H200 GPUs. Throughput is in env/s; step, rollout, and actor times are seconds per training iteration, not milliseconds per action. Ratios denote rollout-side:actor GPU allocation.}
    \label{fig:openvla}
\end{figure}

\textbf{Hardware and models.} 
% All experiments use NVIDIA H200 GPUs with 141\,GB of memory; multi-node runs use an InfiniBand-interconnected cluster. We evaluate $\pi_{0.5}$ (3B parameters), which uses iterative flow matching action generation, and OpenVLA-OFT (7B parameters), which uses autoregressive prediction. Both use action chunking. 
All experiments are conducted on NVIDIA H200 GPUs, each with 141\,GB of memory. Multi-node experiments run on a cluster with InfiniBand interconnects. We evaluate two VLA models: $\pi_{0.5}$ (3B parameters), which generates action chunks via iterative flow matching, and OpenVLA-OFT (7B parameters), which uses parallel decoding with continuous action regression. Figures~\ref{fig:pi_data} and~\ref{fig:openvla} report 8-GPU experiments, Table~\ref{tab:scale} reports 16-GPU experiments, and Table~\ref{tab:multinode_scaling} extends to 256 GPUs.

\textbf{Benchmarks and algorithms.} ManiSkill uses GPU-accelerated physics and rendering, creating competition with model inference for device memory and compute. LIBERO is CPU-intensive and stresses a different execution path. The system benchmarks use GRPO; Section~\ref{sec:algorithms} additionally compares GRPO and PPO on ManiSkill PutOnPlateInScene25Main-v3.

\textbf{Baselines and placement.} We compare RLinf-VLA~\cite{zang2025rlinf} in colocated (RLinf-co), disaggregated (RLinf-dis), and hybrid (RLinf-hyper) modes, and the asynchronous RL-VLA$^3$ framework~\cite{guan2026rl}. The evaluated RL-VLA$^3$ deployment separates environment, rollout, and actor components, whereas \method co-locates environment and rollout inference and isolates actor training. 
% Ratios denote rollout-side:actor GPU allocation: on 8 GPUs, 1:1 corresponds to 4:4 and 3:1 to 6:2; the same proportions apply at larger GPU counts. 
Ratios denote the GPU allocation between rollout inference and policy training. With 8 GPUs, 1:1 assigns 4 GPUs to rollout and 4 to training, while 3:1 assigns 6 GPUs to rollout and 2 to training. The same proportions apply at larger GPU counts.
The ratios describe the nominal resource split, not identical internal placements across frameworks. The hybrid configuration uses two simulator sub-stages (\texttt{pipe=2}); this pipeline-depth setting is aligned across the compared pipelined implementations. Ratios are selected per experiment, not adjusted online.

\textbf{Workload and statistics.} The 16-GPU comparison uses the reported settings of 128 environments/GPU, four rollout epochs, and a global training batch size of 2,048. LIBERO uses a global batch size of 1,024 and a mini-batch size of 128; its per-placement environment counts appear in Table~\ref{tab:libero}. Shared rollout and training settings are matched within each comparison. Throughput is measured in env/s (environment state transitions per second), not action-chunk inference calls per second. Step, rollout, and actor times are seconds per training iteration; step time denotes end-to-end iteration time, not single-action latency. The 16-GPU and LIBERO tables report means and standard deviations across six random seeds; LIBERO explicitly uses sample standard deviation. Statistics for the algorithm comparison have a separate definition in Table~\ref{tab:algorithms}. Appendix~\ref{app:protocol} provides additional configuration details.

\subsection{ManiSkill Throughput and Resource Balance}

\begin{figure}[t]
    \centering
    \includegraphics[width=1\linewidth]{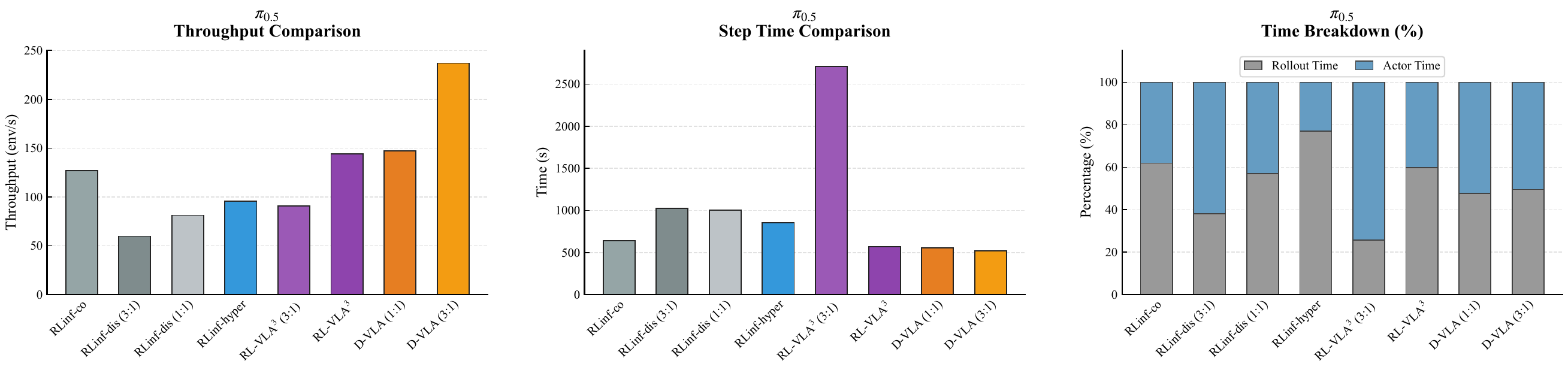}
    \caption{$\pi_{0.5}$ on ManiSkill with 8 H200 GPUs. Left: throughput (env/s). Middle: end-to-end step time (seconds per training iteration). Right: rollout/actor timing breakdown; absolute times are seconds per iteration. Ratios denote rollout-side:actor GPU allocation.}
    \label{fig:pi_data}
\end{figure}

% \textbf{8-GPU and 16-GPU results.} In the 8-GPU $\pi_{0.5}$ experiment (Figure~\ref{fig:pi_data}), \method (1:1) reaches 147.0\,env/s versus 127.24\,env/s for RLinf-co, a 15.53\% improvement. The 3:1 configuration reaches 237.0\,env/s, an 86.26\% improvement over that baseline.  16-GPU results in Table~\ref{tab:scale}, where \method (1:1) reaches 336.04\,env/s. In the 8-GPU OpenVLA-OFT experiment, 156.0\,env/s corresponds to gains of 44.12\% over RLinf-co (108.24\,env/s) and 40.69\% over RL-VLA$^3$ (110.88\,env/s).

\textbf{8-GPU and 16-GPU results.}
For $\pi_{0.5}$ on 8 GPUs (Figure~\ref{fig:pi_data}), \method (1:1) achieves 147.0\,env/s, outperforming
RLinf-co (127.24\,env/s) by 15.53\%. \method (3:1) further increases throughput to 237.0\,env/s, an 86.26\% improvement over RLinf-co. On 16 GPUs, \method (1:1) achieves 336.04\,env/s (Table~\ref{tab:scale}).
For OpenVLA-OFT on 8 GPUs, \method achieves 156.0\,env/s, outperforming RLinf-co (108.24\,env/s) and RL-VLA$^3$ (110.88\,env/s) by 44.12\% and 40.69\%, respectively.

\textbf{Iteration latency and overlap.} The 8-GPU $\pi_{0.5}$ results report a \method iteration time of 566.41\,s versus 1,006.8\,s for RLinf-dis, a 43.74\% reduction. Asynchronous execution overlaps rollout and training rather than adding their durations sequentially. In the OpenVLA-OFT 3:1 configuration, actor-side model updates become the bottleneck. Changing the configured resource ratio toward 1:1 allocates more capacity to training and improves overlap.

\begin{table}[t]
    \centering
    \caption{ManiSkill system comparison on 16 GPUs. Throughput is mean $\pm$ standard deviation over six seeds (env/s). The highest throughput for each model is bold. Full iteration, rollout, and actor timings appear in Table~\ref{tab:timings}.}
    \label{tab:scale}
    \begin{tabular}{lcc}
        \toprule
        Framework & $\pi_{0.5}$ & OpenVLA-OFT \\
        \midrule
        RLinf-co & $232.23 \pm 2.32$ & $87.20 \pm 0.87$ \\
        RLinf-dis (1:1) & $150.58 \pm 1.51$ & $107.23 \pm 1.07$ \\
        RLinf-dis (3:1) & $175.29 \pm 1.75$ & $99.33 \pm 4.99$ \\
        RLinf-hyper & $171.74 \pm 1.72$ & $77.23 \pm 2.77$ \\
        RL-VLA$^3$ (1:1) & $244.61 \pm 2.45$ & $170.48 \pm 2.70$ \\
        RL-VLA$^3$ (3:1) & $250.77 \pm 2.51$ & $152.98 \pm 3.53$ \\
        \method (1:1) & $336.04 \pm 3.36$ & $\mathbf{250.90 \pm 2.51}$ \\
        \method (3:1) & $\mathbf{376.00 \pm 3.76}$ & $154.23 \pm 1.54$ \\
        \bottomrule
    \end{tabular}
\end{table}

\begin{table}[htbp]
    \centering
    \caption{Detailed ManiSkill timing results on 16 GPUs. All timings are reported as mean $\pm$ standard deviation across six seeds, in seconds per training iteration. Step denotes the end-to-end iteration time. Rollout and actor execution may overlap.}
    \label{tab:timings}
    \begin{tabular}{lccc}
        \toprule
        Framework & Step & Rollout & Actor \\
        \midrule
        \multicolumn{4}{c}{$\pi_{0.5}$ (3B)} \\
        \midrule
        RLinf-co & $705.50\pm8.47$ & $503.50\pm6.04$ & $200.50\pm2.41$ \\
        RLinf-dis (1:1) & $1088.00\pm13.06$ & $672.60\pm8.07$ & $383.60\pm4.60$ \\
        RLinf-dis (3:1) & $2840.00\pm34.08$ & $1662.02\pm19.94$ & $1142.30\pm13.71$ \\
        RLinf-hyper & $954.00\pm11.45$ & $751.20\pm9.01$ & $199.00\pm2.39$ \\
        RL-VLA$^3$ (1:1) & $669.80\pm8.04$ & $525.66\pm6.31$ & $383.00\pm4.60$ \\
        RL-VLA$^3$ (3:1) & $1960.00\pm23.52$ & $1580.24\pm18.96$ & $1420.33\pm17.04$ \\
        \method (1:1) & $488.32\pm5.86$ & $460.65\pm5.53$ & $452.33\pm5.43$ \\
        \method (3:1) & $1307.23\pm15.69$ & $1299.20\pm15.59$ & $1288.02\pm15.46$ \\
        \midrule
        \multicolumn{4}{c}{OpenVLA-OFT (7B)} \\
        \midrule
        RLinf-co & $1877.00\pm22.52$ & $259.00\pm3.11$ & $1607.50\pm19.29$ \\
        RLinf-dis (1:1) & $1528.00\pm18.34$ & $508.52\pm6.10$ & $1001.80\pm12.02$ \\
        RLinf-dis (3:1) & $1628.00\pm59.38$ & $896.60\pm10.76$ & $1030.23\pm48.36$ \\
        RLinf-hyper & $2123.50\pm25.48$ & $432.75\pm5.19$ & $1675.00\pm20.10$ \\
        RL-VLA$^3$ (1:1) & $961.03\pm11.53$ & $354.20\pm4.25$ & $814.40\pm9.77$ \\
        RL-VLA$^3$ (3:1) & $1058.35\pm12.70$ & $761.60\pm9.14$ & $816.40\pm9.80$ \\
        \method (1:1) & $653.23\pm7.84$ & $508.23\pm6.10$ & $568.24\pm6.82$ \\
        \method (3:1) & $1062.30\pm12.75$ & $708.23\pm8.50$ & $868.24\pm10.42$ \\
        \bottomrule
    \end{tabular}
\end{table}

\textbf{Structural comparison with RL-VLA$^3$.} Both frameworks overlap rollout and training, but differ in observation locality. In the tested RL-VLA$^3$ placement, observations cross GPU boundaries between environment and inference workers; \method accesses them within the same process. On 16 GPUs with OpenVLA-OFT and a 3:1 split, throughput is close (154.23 versus 152.98\,env/s). At 1:1, \method reaches 250.90 versus 170.48\,env/s, a 47.17\% improvement. Section~\ref{sec:ablations} reports isolated transfer-latency measurements for the data path.

\textbf{Compare with RLinf-hyper.}  RLinf-hybrid placement is not uniformly preferable in baseline. For OpenVLA-OFT at 16 GPUs, RLinf-hyper is below both RLinf-co and RLinf-dis, while for $\pi_{0.5}$ it exceeds RLinf-dis (1:1). The former ordering agrees with the corresponding 16-GPU, \texttt{pipe=2} ManiSkill configuration reported for RLinf-VLA~\cite{zang2025rlinf}. 
% Our experiments use H200 GPUs, while that report uses A100 GPUs.

\begin{figure}[t]
    \centering
    \includegraphics[width=0.9\linewidth]{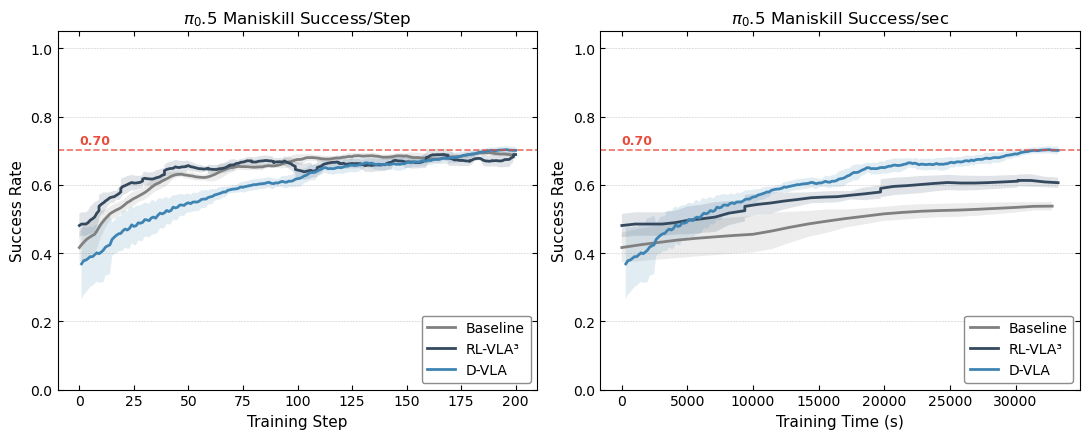}
    \caption{Training success rate on ManiSkill with $\pi_{0.5}$ under the evaluated system configurations.}
    \label{fig:success}
\end{figure}

\textbf{Learning behavior.} The ManiSkill success-rate curves in Figure~\ref{fig:success} show competitive learning behavior under the evaluated asynchronous configuration. Section~\ref{sec:algorithms} reports a separate task-level comparison for both GRPO and PPO.

\subsection{CPU-Intensive LIBERO Evaluation}
\label{sec:libero}

Table~\ref{tab:libero} complements the GPU-intensive ManiSkill experiments with 8-GPU LIBERO measurements. \method achieves the highest throughput in every row,
with gains over RLinf ranging from 20.83\% to 76.99\%. On this CPU-intensive simulator, the source of benefit differs from the GPU-physics contention motivating our design: the asynchronous Swimlane pipeline overlaps rollout with actor updates and reduces scheduling and serialization stalls, rather than resolving the same on-GPU physics contention. 

\begin{table}[t]
    \centering
    \caption{LIBERO steady-state throughput on 8 GPUs (env/s, mean $\pm$ sample standard deviation over six seeds). Ratios 6:2 and 4:4 correspond to 3:1 and 1:1.}
    \label{tab:libero}
    \resizebox{\linewidth}{!}{
    \begin{tabular}{lcccccc}
        \toprule
        Model & Roll:act & Env/GPU & Total env & RLinf & RL-VLA$^3$ & \method \\
        \midrule
        $\pi_{0.5}$ & 6:2 & 32 & 192 & $168.63\pm0.78$ & $227.05\pm2.10$ & $298.45\pm3.41$ \\
        $\pi_{0.5}$ & 6:2 & 64 & 384 & $160.10\pm0.87$ & $208.99\pm2.80$ & $268.75\pm5.22$ \\
        $\pi_{0.5}$ & 4:4 & 32 & 128 & $165.30\pm2.01$ & $184.80\pm2.90$ & $208.63\pm3.64$ \\
        $\pi_{0.5}$ & 4:4 & 64 & 256 & $158.50\pm1.10$ & $173.36\pm2.50$ & $191.52\pm3.72$ \\
        \midrule
        OpenVLA-OFT & 6:2 & 32 & 192 & $161.60\pm0.02$ & $191.66\pm0.40$ & $228.40\pm0.54$ \\
        OpenVLA-OFT & 6:2 & 32 & 128 & $145.40\pm2.75$ & $173.26\pm1.60$ & $207.30\pm0.75$ \\
        OpenVLA-OFT & 4:4 & 32 & 128 & $181.22\pm0.62$ & $214.11\pm0.50$ & $254.30\pm0.12$ \\
        OpenVLA-OFT & 4:4 & 64 & 256 & $174.58\pm0.39$ & $196.99\pm0.40$ & $224.38\pm0.45$ \\
        \bottomrule
    \end{tabular}}
\end{table}

\subsection{Mechanism-Level Ablations}
\label{sec:ablations}

We evaluate three mechanisms separately: the weight-sync backend for synchronization stability, dual-pool memory management for maximum stable environment concurrency, and in-process observation access for transfer latency. Appendix~\ref{app:ablations} provides the complete configurations and measurements.

\textbf{Weight-sync backend.} Replacing Gloo with NCCL in the tested $\pi_{0.5}$ production configuration produces a first-step deadlock at \texttt{cudaDeviceSynchronize} when communication shares the reset synchronization path. The process times out with exit code 124 before completing the first step. With Gloo, training completes with exit code 0; the delayed-peer synchronization probe takes about 0.1\,ms, and synchronization overhead before \texttt{env.reset} is approximately 0.02--0.16\,ms per step. The result establishes stability for this schedule.

\textbf{Dual-pool memory.} With partitioning enabled, Torch-reserved memory at rollout end decreases from 39,286 to 33,388\,MiB, and free device memory increases from 19,588 to 25,486\,MiB (about 5.8\,GiB). Maximum stable concurrency increases from 4,352 to 4,608 environments, or from 544 to 576 per GPU. Without partitioning, 4,480 environments fail during PhysX buffer creation; with partitioning, 4,864 environments reach a PyTorch out-of-memory limit. Thus the partition raises the capacity ceiling.

\textbf{In-process observation access.} For batches of 32, 64, and 128 observations (4.6, 9.2, and 18.4\,MiB), zero-copy access takes 7.26, 7.48, and 7.51\,$\mu$s, compared with 691, 697, and 706\,$\mu$s for the measured NCCL transfer path. The zero-copy path costs approximately 1.1\% of the transfer latency and changes little with batch size. These measurements quantify observation-access latency in the co-located data path.

\subsection{GPU-Count Scaling and Environment-Count Saturation}
\label{sec:scaling}

\textbf{Multi-node scaling.} Table~\ref{tab:multinode_scaling} evaluates $\pi_{0.5}$ with GRPO and a 3:1 split from 8 to 256 GPUs. The reported environment setting is 192 per GPU, the global batch size is $192G$ for $G$ total GPUs, and the mini-batch size is 192. Scaling efficiency relative to 8 GPUs is defined in
Equation~\ref{eq:scaling_efficiency}. With a $32\times$ increase in GPU count, \method retains 86.70\% efficiency, compared with 73.16\% for RL-VLA$^3$ and 62.02\% for RLinf. The evaluated systems differ in communication locality and scheduling: \method keeps environment--rollout traffic within each node and synchronizes the model across actor workers; the fully disaggregated RL-VLA$^3$ layout permits observation traffic across nodes, while synchronous RLinf exposes stage waiting.

\begin{equation}
    \operatorname{Eff}(G)=
    \frac{\operatorname{Thr}(G)}{\operatorname{Thr}(8)\,(G/8)}\times100\%.
    \label{eq:scaling_efficiency}
\end{equation}

\begin{table}[t]
    \centering
    \caption{$\pi_{0.5}$  scaling on ManiSkill with GRPO, a 3:1 split, and the reported 192-environment/GPU budget. Throughput (Thr.) is in env/s. }
    \label{tab:multinode_scaling}
    \resizebox{\linewidth}{!}{
    \begin{tabular}{rrcccccc}
        \toprule
        GPUs & Global batch & \multicolumn{2}{c}{\method} & \multicolumn{2}{c}{RL-VLA$^3$} & \multicolumn{2}{c}{RLinf} \\
        \cmidrule(lr){3-4}\cmidrule(lr){5-6}\cmidrule(lr){7-8}
        & & Thr. & Eff. & Thr. & Eff. & Thr. & Eff. \\
        \midrule
        8 & 1536 & 192.94 & 100.00\% & 130.64 & 100.00\% & 94.03 & 100.00\% \\
        16 & 3072 & 376.23 & 97.50\% & 248.22 & 95.00\% & 174.07 & 92.56\% \\
        32 & 6144 & 714.84 & 92.62\% & 446.80 & 85.50\% & 296.34 & 78.79\% \\
        64 & 12288 & 1406.65 & 91.13\% & 854.27 & 81.74\% & 550.20 & 73.14\% \\
        128 & 24576 & 2720.29 & 88.12\% & 1608.27 & 76.94\% & 1002.92 & 66.66\% \\
        256 & 49152 & 5352.70 & 86.70\% & 3058.47 & 73.16\% & 1866.03 & 62.02\% \\
        \bottomrule
    \end{tabular}}
\end{table}

\begin{figure}[t]
    \centering
    \includegraphics[width=1\linewidth]{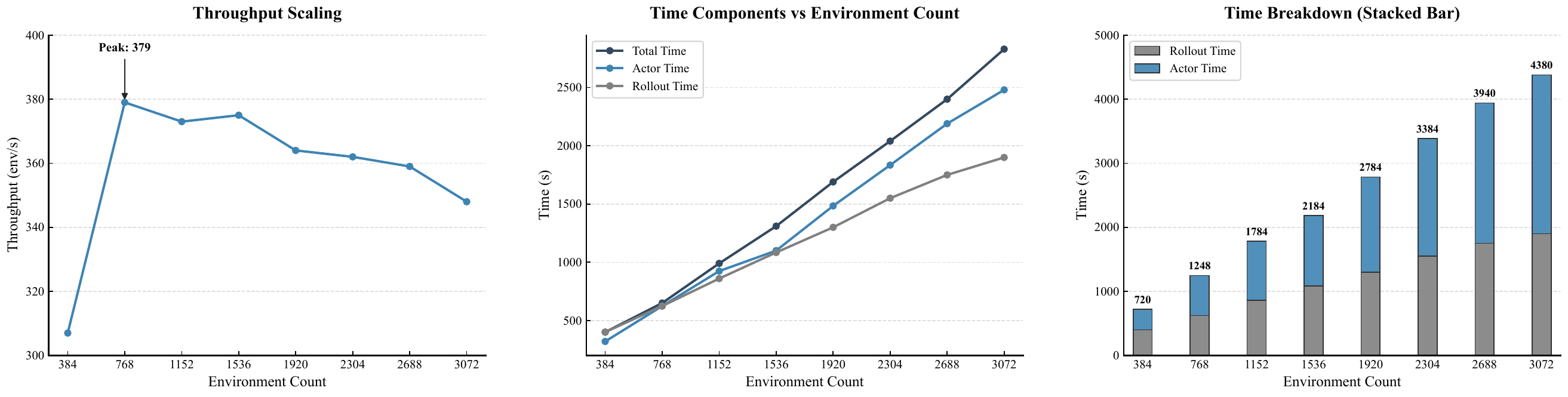}
    \caption{Fixed-hardware environment-count sweep for $\pi_{0.5}$ with a 3:1 rollout-side:actor allocation. Throughput is in env/s; step, rollout, and actor times are seconds per training iteration. 
    % This is not the multi-node scaling study in Table~\ref{tab:multinode_scaling}.
    }
    \label{fig:scale}
\end{figure}

\textbf{Environment-count operating range on fixed hardware.} A separate $\pi_{0.5}$ sweep increases environments from 384 to 3,072 while holding hardware and the 3:1 resource allocation fixed (Figure~\ref{fig:scale}). Throughput peaks near 379\,env/s at 768 environments, then decreases toward 360\,env/s at 3,072. Around the peak, rollout and actor execution times are approximately balanced under the fixed GPU allocation. The sweep identifies the throughput peak for this configuration and shows saturation as the environment count increases beyond it.

\subsection{GRPO and PPO Comparison}
\label{sec:algorithms}

We compare both objectives on ManiSkill PutOnPlateInScene25Main-v3 using $\pi_{0.5}$, 384 training environments, a global batch size of 2,048, and 300 training iterations. Runs start from the same SFT checkpoint, with shared model and optimization settings aligned across frameworks. 
% Evaluation and checkpointing are disabled during training.
Appendix~\ref{app:algorithms} lists the shared settings and objective-specific differences.

\begin{table}[h!]
    \centering
    \caption{GRPO/PPO comparison on ManiSkill PutOnPlateInScene25Main-v3 with $\pi_{0.5}$. Throughput is reported as mean $\pm$ standard deviation (env/s). Success is training-trajectory success, with mean $\pm$ standard deviation over the final 50 training steps in the reported converged segment near steps 250--300.}
    \label{tab:algorithms}
    \begin{tabular}{llcc}
        \toprule
        Algorithm & Framework & Throughput $\uparrow$ & Success (\%) $\uparrow$ \\
        \midrule
        GRPO & RLinf-co & $230.3\pm2.1$ & $58.84\pm5.04$ \\
        GRPO & RL-VLA$^3$ & $264.6\pm2.9$ & $63.77\pm3.58$ \\
        GRPO & \method & $306.8\pm3.7$ & $78.69\pm2.12$ \\
        \midrule
        PPO & RLinf-co & $175.4\pm1.7$ & $73.66\pm1.68$ \\
        PPO & RL-VLA$^3$ & $179.5\pm1.3$ & $80.00\pm1.59$ \\
        PPO & \method & $183.6\pm0.8$ & $86.34\pm1.50$ \\
        \bottomrule
    \end{tabular}
\end{table}

Under GRPO, \method improves throughput by 33.22\% over RLinf-co and training success by 19.85 percentage points. Under PPO, the corresponding improvements are 4.68\% and 12.68 percentage points. These results demonstrate throughput and training-success improvements with both GRPO and PPO in \method.

\section{Conclusion and Future Work}
\label{sec:con}

\method separates trajectory transfer and weight synchronization, overlaps execution with a four-thread pipeline, and combines memory partitioning with local observation access. Experiments with 3B and 7B VLA models cover GPU-intensive ManiSkill and CPU-intensive LIBERO. The targeted ablations demonstrate synchronization stability, increased memory capacity, and reduced transfer latency; end-to-end experiments demonstrate throughput benefits with GRPO and PPO. A separate $\pi_{0.5}$ scaling experiment retains 86.70\% efficiency at 256 GPUs relative to 8 GPUs.

\textbf{Future work.} We plan to extend \method to multi-agent interaction and additional physics engines, and evaluate its behavior under a broader range of policy staleness.

\section*{Acknowledgement}
This work is supported by the Natural Science Foundation of Guangdong Province (Grant No. 2025A1515011946)

\bibliography{mybib}
\bibliographystyle{unsrt}

\appendix

\section{Experimental Protocol }
\label{app:protocol}

\textbf{Reporting scope.} The 16-GPU ManiSkill and 8-GPU LIBERO measurements report means and standard deviations across six seeds. LIBERO explicitly reports sample standard deviation: for measurements $x_1,\ldots,x_6$, $\bar{x}=\sum_{i=1}^{6}x_i/6$ and $s=\sqrt{\sum_{i=1}^{6}(x_i-\bar{x})^2/5}$. 

\textbf{Hardware configuration.} The reported hardware is H200 (141\,GB) with InfiniBand for inter-node communication.

\section{Mechanism-Level Measurement Details}
\label{app:ablations}

\subsection{Weight Synchronization Stability}

Table~\ref{tab:backend_ablation} replaces the weight-synchronization backend while retaining the tested production scheduling path for $\pi_{0.5}$. With NCCL, the shared synchronization path creates a circular dependency between weight synchronization and rollout-data progress at the first reset. The NCCL run fails with a timeout before completing the first step. With Gloo, the host-side communication thread can progress without waiting on the same CUDA path.

\begin{table}[htbp]
    \centering
    \caption{Weight-sync backend comparison in the tested $\pi_{0.5}$ configuration. The probe measures \texttt{cudaDeviceSynchronize} with delayed peer progress; pre-reset overhead is synchronization time before \texttt{env.reset}.}
    \label{tab:backend_ablation}
    \begin{tabular}{lccc}
        \toprule
        Backend & Probe & Exit code & Pre-reset overhead \\
        \midrule
        Gloo & $\sim$0.1\,ms & 0 (completes) & 0.02--0.16\,ms/step \\
        NCCL & Deadlock & 124 (timeout) & No completed step \\
        \bottomrule
    \end{tabular}
\end{table}

\subsection{Memory Capacity}

The memory experiment targets 4,608 training environments (576 per GPU) and 1,024 evaluation environments (128 per GPU), with \path{predict_mbs=64}, \path{force_lazy_eval=true}, and an ON-phase fraction of 0.24. Table~\ref{tab:memory_ablation} reports Torch-reserved memory, free device memory, and exit status for the target configuration, alongside the maximum stable concurrency from the capacity sweep. Torch-reserved memory measures allocator reservation.

\begin{table}[htbp]
    \centering
    \caption{Dual-pool partitioning. Memory readings are at rollout end (MiB). Exit code is for the target 4,608-environment configuration; maximum stable concurrency and the next reported failing configuration come from the capacity sweep.}
    \label{tab:memory_ablation}
    \resizebox{\linewidth}{!}{
    \begin{tabular}{lcccccl}
        \toprule
        Partition & Exit code & Torch reserved & Free & Max env & Env/GPU & Reported failure beyond capacity \\
        \midrule
        OFF & 1 & 39286 & 19588 & 4352 & 544 & 4480: PhysX buffer creation \\
        ON & 0 & 33388 & 25486 & 4608 & 576 & 4864: PyTorch OOM \\
        \bottomrule
    \end{tabular}}
\end{table}

\subsection{Observation Transfer Latency}

Table~\ref{tab:zerocopy_ablation} measures observation access within a co-located process against the tested NCCL transfer path. The former avoids transporting the observation payload between separate components. These measurements describe the complete measured paths, including their access/transfer overhead.

\begin{table}[htbp]
    \centering
    \caption{Observed latency of the in-process zero-copy and NCCL observation paths. Ratios are computed from the reported latencies.}
    \label{tab:zerocopy_ablation}
    \begin{tabular}{rcccc}
        \toprule
        Batch & Observation (MiB) & Zero-copy ($\mu$s) & NCCL ($\mu$s) & Ratio \\
        \midrule
        32 & 4.6 & 7.26 & 691 & 1.05\% \\
        64 & 9.2 & 7.48 & 697 & 1.07\% \\
        128 & 18.4 & 7.51 & 706 & 1.06\% \\
        \bottomrule
    \end{tabular}
\end{table}

\section{GRPO and PPO Configuration}
\label{app:algorithms}

The task is ManiSkill PutOnPlateInScene25Main-v3 with GPU simulation and RGB plus segmentation observations. Shared settings are \path{max_episode_steps=80}, \path{num_action_chunks=5} (16 chunk samples per trajectory), and the \texttt{pi05} model configured with \path{flow_noise}, \path{noise_params=[0.16,0.12,200]}, and \path{num_steps=4}. All runs start from the same SFT checkpoint, use 300 training steps, and disable evaluation and checkpointing during training. The policy optimizer uses a learning rate of $7.91\times10^{-6}$, Adam betas $(0.9,0.95)$, and gradient clipping at 1.0; global batch size is 2,048.
Table~\ref{tab:algorithm_batches} summarizes the batch configurations for RLinf and \method. 

\begin{table}[htbp]
    \centering
    \caption{Reported environment and batch settings for the GRPO/PPO comparison. Objective-specific differences are kept the same between RLinf and \method.}
    \label{tab:algorithm_batches}
    \resizebox{\linewidth}{!}{
    \begin{tabular}{lcccc}
        \toprule
        Setting & RLinf GRPO & RLinf PPO & \method GRPO & \method PPO \\
        \midrule
        Training environments & 384 & 384 & 384 & 384 \\
        Evaluation environments & 128 & 240 & 128 & 240 \\
        Group size & 8 (48 groups) & 1 & 8 (48 groups) & 1 \\
        Micro-batch size & 64 & 64 & 64 & 64 \\
        Update epochs & 1 & 2 & 1 & 2 \\
        \bottomrule
    \end{tabular}}
\end{table}

\begin{table}[htbp]
    \centering
    \caption{Objective-specific hyperparameters. These differ between GRPO and PPO, rather than between the frameworks within an objective.}
    \label{tab:algorithm_hyperparameters}
    \begin{tabular}{lcc}
        \toprule
        Parameter & GRPO & PPO \\
        \midrule
        Advantage estimator & GRPO & GAE ($\gamma=0.99$, $\lambda=0.95$) \\
        Entropy bonus & 0.005 & 0.005 \\
        Policy clipping (lower/upper) & 0.2 / 0.2 & 0.2 / 0.2 \\
        Value clipping & --- & 0.2 \\
        Huber parameter & --- & 10 \\
        Reward filtering & True, $[0.0,1.0]$ & False \\
        Value learning rate & --- & $1.55\times10^{-4}$ \\
        Adam epsilon & $10^{-8}$ & $10^{-5}$ \\
        \bottomrule
    \end{tabular}
\end{table}

\newpage

\end{document}